\pdfoutput=1

\documentclass[11pt]{article}
\usepackage{CJKutf8}
\usepackage{acl}

\usepackage{times}
\usepackage{latexsym}

\usepackage[T1]{fontenc}

\usepackage[utf8]{inputenc}

\usepackage{microtype}

\usepackage{inconsolata}

\usepackage{times}
\usepackage{latexsym}
\usepackage{balance}
\usepackage[T1]{fontenc}

\usepackage[utf8]{inputenc}

\usepackage{microtype}

\usepackage{inconsolata}
\usepackage{CJKutf8}
\usepackage{booktabs}
\usepackage{graphicx}
\usepackage{array}
\usepackage{bm}
\usepackage{amsmath}
\usepackage{enumitem}
\usepackage{threeparttable}
\usepackage{multirow}
\usepackage{makecell}

\usepackage{latexsym} 

\usepackage[T1]{fontenc}

\usepackage[utf8]{inputenc}
\usepackage{float}

\usepackage{microtype}

\usepackage{xcolor}

\def\BibTeX{{\rm B\kern-.05em{\sc i\kern-.025em b}\kern-.08em
    T\kern-.1667em\lower.7ex\hbox{E}\kern-.125emX}}
\usepackage{cite}
\usepackage{amsmath,amssymb,amsfonts}
\usepackage{algorithmic}
\usepackage[ruled,linesnumbered,noline,noend]{algorithm2e}

\SetCommentSty{mycommfont}

\usepackage[medium,compact]{titlesec}

\SetNlSty{}{}{}

\let\oldnl\nl
\newcommand\nonl{%
  \renewcommand{\nl}{\let\nl\oldnl}}
  
\usepackage{hyperref}
\hypersetup{
    colorlinks=true,
    linkcolor=blue,}
\AtBeginDocument{%
  \providecommand\BibTeX{{%
    \normalfont B\kern-0.5em{\scshape i\kern-0.25em b}\kern-0.8em\TeX}}}

\usepackage{textcomp}
\usepackage{xcolor}
\def\BibTeX{{\rm B\kern-.05em{\sc i\kern-.025em b}\kern-.08em
    T\kern-.1667em\lower.7ex\hbox{E}\kern-.125emX}}
\usepackage{amsmath,amssymb,amsfonts}
\usepackage{algorithmic}
\usepackage[ruled,linesnumbered,noline,noend]{algorithm2e}
\usepackage{threeparttable}
\usepackage{booktabs}
\usepackage{graphicx}
\usepackage{textcomp}
\usepackage{array}
\usepackage{bm}
\usepackage{amsmath}
\usepackage{color, colortbl}
\SetCommentSty{mycommfont}

\usepackage[misc]{ifsym}
\usepackage{amsthm}
\newtheorem{Problem definition}{Problem definition}

\def\BibTeX{{\rm B\kern-.05em{\sc i\kern-.025em b}\kern-.08em
    T\kern-.1667em\lower.7ex\hbox{E}\kern-.125emX}}
\usepackage{hyperref}
\hypersetup{
    colorlinks=true,
    linkcolor=blue,}

\usepackage{color, colortbl}
\usepackage{array}
\usepackage{bm}
\usepackage{amsmath}
\usepackage{threeparttable}
\usepackage{booktabs}
\usepackage{graphicx}
\usepackage{hyperref}
\usepackage{multirow}
\usepackage[ruled,linesnumbered,noline,noend]{algorithm2e}
\usepackage{makecell}
\hypersetup{
    colorlinks=true,
    linkcolor=blue,}
\usepackage{CJKutf8}
\usepackage{booktabs}
\usepackage{graphicx}
\usepackage{textcomp}
\usepackage{array}
\usepackage{bm}
\usepackage{amsmath}

\usepackage{threeparttable}
\usepackage{multirow}
\usepackage{makecell}
\usepackage{times}
\usepackage{latexsym}
\usepackage[T1]{fontenc}

\usepackage[utf8]{inputenc}
\usepackage{float}

\usepackage{microtype}

\usepackage{xcolor}

\def\BibTeX{{\rm B\kern-.05em{\sc i\kern-.025em b}\kern-.08em
    T\kern-.1667em\lower.7ex\hbox{E}\kern-.125emX}}
\usepackage{cite}
\usepackage{amsmath,amssymb,amsfonts}
\usepackage{algorithmic}
\usepackage[ruled,linesnumbered,noline,noend]{algorithm2e}

\SetCommentSty{mycommfont}

\usepackage{fontawesome}

\SetNlSty{}{}{}

\let\oldnl\nl

\usepackage{hyperref}
\hypersetup{
    colorlinks=true,
    linkcolor=blue,}
\AtBeginDocument{%
  \providecommand\BibTeX{{%
    \normalfont B\kern-0.5em{\scshape i\kern-0.25em b}\kern-0.8em\TeX}}}
\usepackage[medium,compact]{titlesec}

\usepackage{times}
\usepackage{latexsym}

\usepackage[T1]{fontenc}

\usepackage[utf8]{inputenc}

\usepackage{microtype}
\usepackage{mathrsfs}

\title{Do Large Language Models have Problem-Solving Capability under Incomplete Information Scenarios?}

\author{Yuyan Chen$^{1}$, Tianhao Yu$^{2}$, Yueze Li$^{1}$, Songzhou Yan$^{1}$, Sijia Liu$^{1}$, \\ \textbf{Jiaqing Liang}$^{3}$, \textbf{Yanghua Xiao}$^{1}$ $^{\textrm{\Letter}}$\\
        $^1$Shanghai Key Laboratory of Data Science, School of Computer Science, Fudan University, \\
        $^2$SenseAuto,
        $^3$School of Data Science, Fudan University\\
        \texttt{\{chenyuyan21, yuezeli23, szyan21, sijialiu21\}@m.fudan.edu.cn},\\
        \texttt{\{liangjiaqing, shawyh\}@fudan.edu.cn},  \texttt{avery1026@hotmail.com} 
        }

\begin{document}
\maketitle
\begin{abstract}
The evaluation of the problem-solving capability under incomplete information scenarios of Large Language Models (LLMs) is increasingly important, encompassing capabilities such as questioning, knowledge search, error detection, and path planning. Current research mainly focus on LLMs' problem-solving capability such as ``Twenty Questions''.
However, these kinds of games do not require recognizing misleading cues which are necessary in the incomplete information scenario.
Moreover, the existing game such as 
``Who is undercover'' are highly subjective, making it challenging for evaluation.
Therefore, in this paper, we introduce a novel game named BrainKing based on the ``Who is undercover'' and ``Twenty Questions'' for evaluating LLM capabilities under incomplete information scenarios. It requires LLMs to identify target entities with limited yes-or-no questions and potential misleading answers. By setting up easy, medium, and hard difficulty modes, we comprehensively assess the performance of LLMs across various aspects. Our results reveal the capabilities and limitations of LLMs in BrainKing, providing significant insights of LLM problem-solving levels.
\end{abstract}

\section{Introduction}

Incomplete information scenarios include missing information, uncertainty, and misinformation, encountered in fields such as business negotiations, military strategy, medical diagnosis, and legal judgments~\citep{chen2024recent, gibbons1992primer, chen2024temporalmed,neo2024towards,jin2024agentreview}. 
The problem-solving capability under incomplete information refer to the capability to effectively handle available information, make rational inferences, and decisions in situations lacking comprehensive data. This capability is crucial in real life as we cannot possess all the necessary information for decision-making~\citep{chen2023xmqas,chen2024xmecap,chen2024hotvcom,chen2022grow}. 
It's also important for large language models (LLMs), which not only tests LLMs' logical reasoning capabilities but also involves adjusting strategies in constantly changing environments, significantly enhancing their robustness and quality of decision-making in various fields. Therefore, a natural question arises: \emph{Do LLMs have problem-solving capability under incomplete information scenarios?}

\begin{figure}[t]
  \centering
  \includegraphics[width=0.9\linewidth]{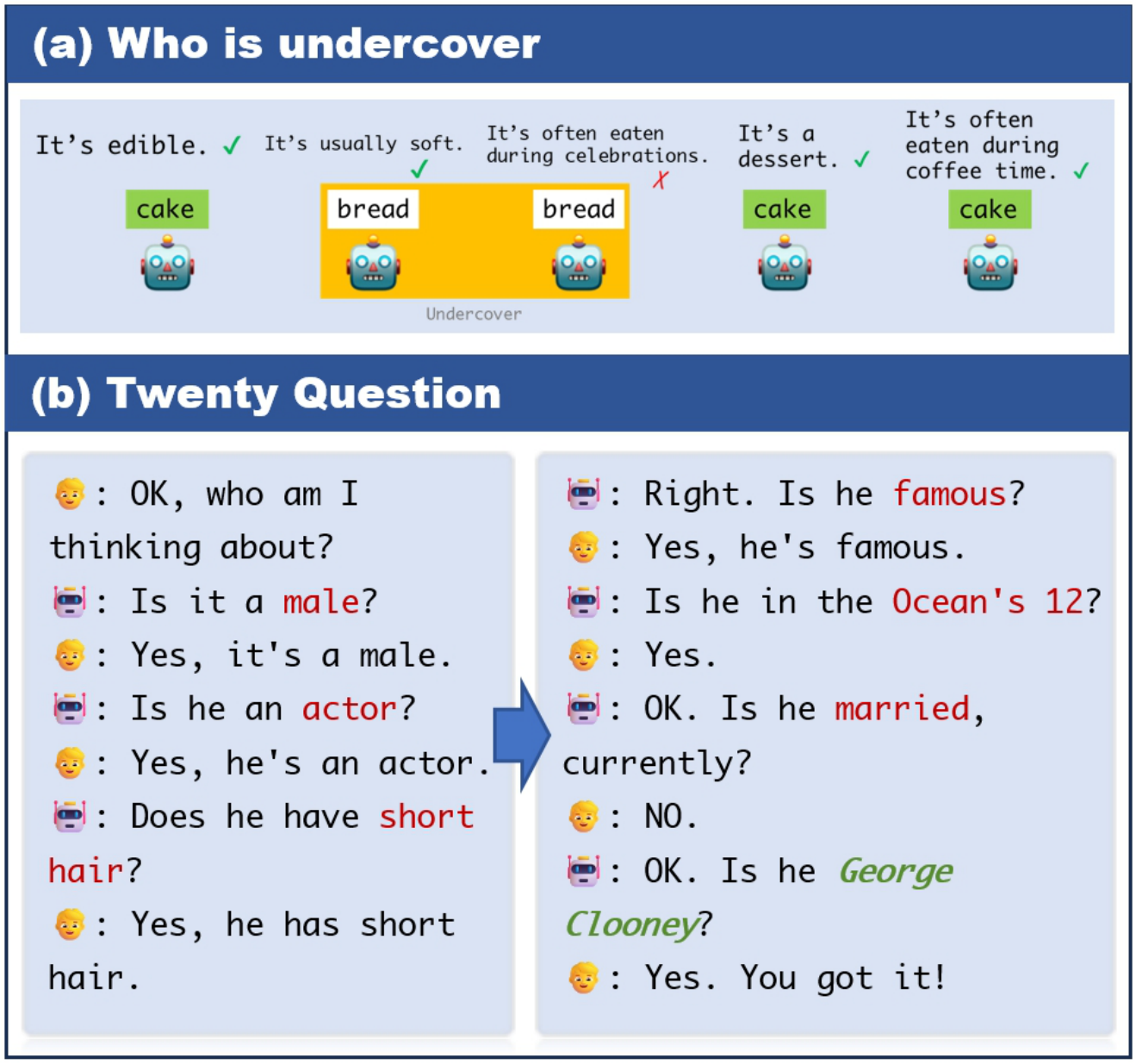}
  \caption{A sample of the ``Who is undercover'' game (a) and the ``Twenty Questions'' game (b).}
  \label{fig:zhishang-intro}
\end{figure}

Previous research on problem-solving capability under incomplete information scenarios focus on simulating complex decision-making environments through games~\citep{jin2022towards,yang2022reinforcement,zhao2023competeai,jin2023predicting}, such as ``Werewolf''~\citep{xu2023exploring, ri2022dynamics, toriumi2017ai}, ``Poker''~\citep{brown2019superhuman} and ``Avalon''~\citep{light2023avalonbench}, etc. These games require players to make decisions without full information, often involving deception and strategic planning to conceal their real identities. 
As shown in Fig.~\ref{fig:zhishang-intro}(a), ``Who is undercover'' is another incomplete information game which requires players to deduce whether they are the spy based on others' descriptions. In this running example, player with ``bread'' is the spy in this game and he needs to hide himself against being caught through twist the facts in his description like ``It typically requires more sugar, fats, and eggs.''.
However, even advanced LLMs like GPT-3 and GPT-3.5, while excelling in general NLP tasks~\citep{tao2024nevlp, liu2023reta, xiong2024largelanguagemodelslearn, chung2023increasing,chen2024talk}, revealing limitations in effective decision-making under incomplete information environments~\citep{gigerenzer2011heuristic, binz2023using,chen2024emotionqueen,chen2024drAcademy,ni2024timeseries,ni2024earnings,202407.0981}.

Information processing is a crucial problem-solving capability under incomplete information scenarios, exemplified by the Minesweeper game~\citep{li2023assessing}, the Twenty Questions game~\citep{walsorth1882twenty, giordano1998using}, etc.
As shown in Fig.~\ref{fig:zhishang-intro}(b), the Twenty Questions game requests the player to pose a series of yes-or-no questions to guess the given entity (i.e. ``George clooney''), which can effectively evaluate LLMs' creativity~\citep{hu2018playing}, knowledge retrieval~\citep{williams2015twenty, szymanski2012information,chen2023can,202407.2102,zhou2024reconstruction}, multi-hop reasoning capabilities~\citep{noever2023chatbots, siegler1977twenty,chen2023mapo}. 
However, the above-mentioned games, such as Twenty Questions, do not adequately assess LLMs' capabilities in processing information and solving problems because it lacks deception and strategic complexity that require recognizing misleading cues and formulating adaptive strategies based on limited or false information.
Moreover, games like ``Werewolf'' and ``Who is undercover'' are highly subjective, making it challenging to evaluate LLMs' capabilities under incomplete information scenarios effectively.

Therefore, in this paper, we introduce a new game named BrainKing by combining the ``Who is undercover'' and ``Twenty Questions'' game to assess LLMs' information processing and problem-solving capabilities under incomplete information scenarios. BrainKing challenges LLM participants to identify entities amidst potential misinformation through a limited set of yes-or-no questions across easy, medium, and hard difficulty modes. This game objectively assesses LLMs' world knowledge, reverse thinking, and error detection capabilities. Our results explore five key questions regarding LLM performance in BrainKing, investigating the relationship between accuracy and rounds, the impact of starting point difficulty and the number of wrong answers, and the correlation between accuracy and the ability to recognize confusion. This assessment, including questioning, knowledge retrieval, misinformation recognition, and planning abilities, effectively evaluates LLMs' capabilities under incomplete information scenarios through a simple-yet-effective game with objective metrics.
In summary, our research makes three key contributions:
\begin{itemize}
    \item We propose a simple-yet-effective game named BrainKing to evaluate LLMs' information processing and problem-solving capabilities under incomplete information scenarios.
    \item Based on the BrainKing game, we introduce an automated evaluation methodology that objectively measures LLMs' performance.
    \item We conduct experiments to comprehensively assess existing LLMs' capabilities under incomplete information scenarios, yielding significant conclusions for enhancing their capabilities and limitations.
\end{itemize}

\begin{figure*}[!h]
  \centering
  \includegraphics[width=0.8\linewidth]{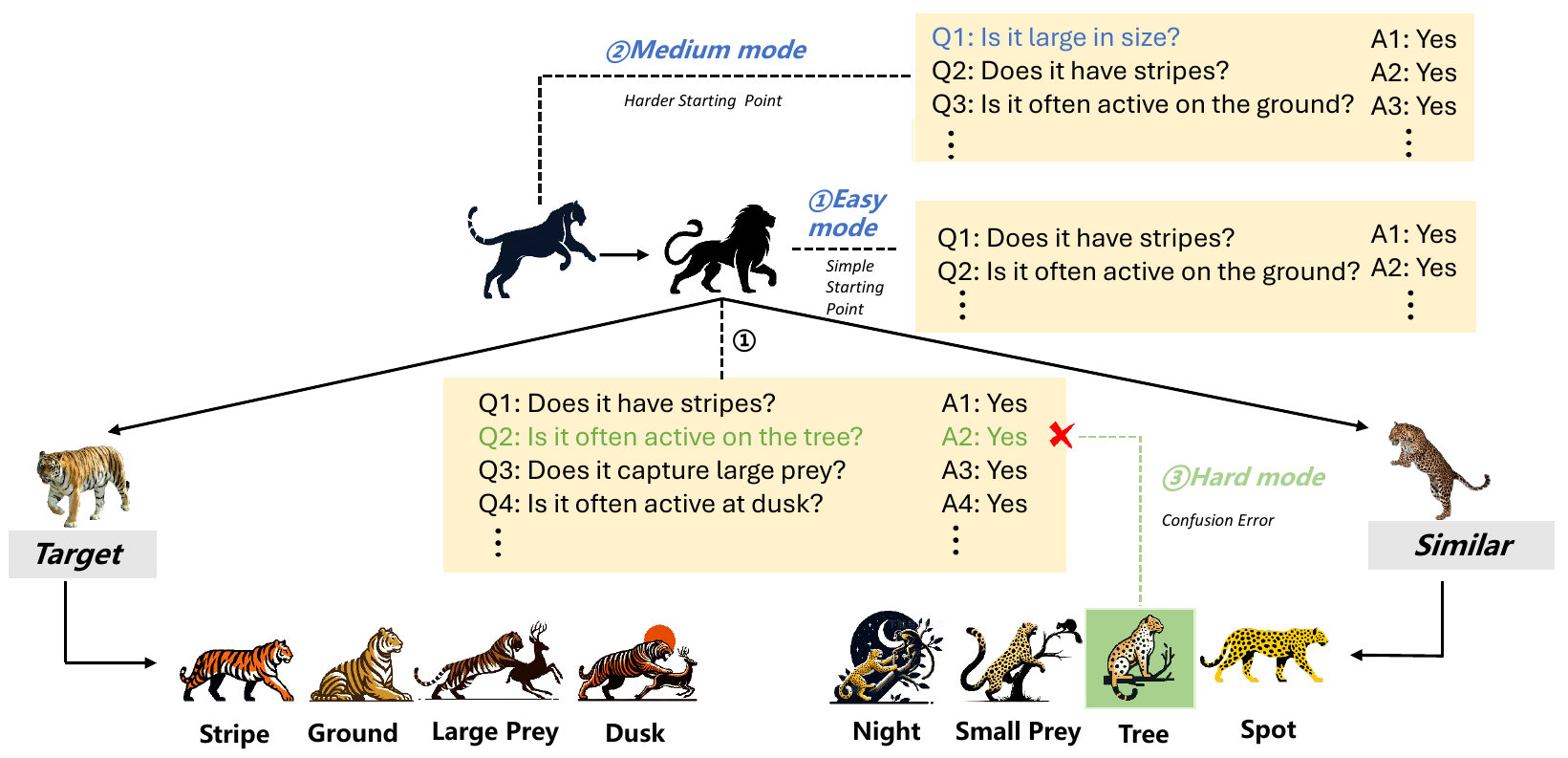}
  \caption{The overview of the proposed BrainKing benchmark, including three modes.}
  \label{fig:zhishang-framework}
\end{figure*}

\section{Datasets and Task Setups}






The proposed BrainKing game, inspired by the Twenty Questions game, requires LLM participants to identify an entity with a limited number of yes-or-no questions despite potentially misleading answers. It contains three difficulty modes: easy, medium, and hard, as illustrated in Fig.~\ref{fig:zhishang-framework}. These modes are designed to thoroughly assess an LLM's world knowledge, reverse thinking, and error detection capabilities in identifying the target entity.

Specifically, we utilize an open-source Twenty Questions dataset~\footnote{https://github.com/allenai/twentyquestions}, which comprises 78,890 entities. To ensure the task is manageable and not hindered by a lack of knowledge by all LLM participants, we adopt GPT3.5~\footnote{https://chat.openai.com/} to score each entity for commonness from 1 (less common) to 3 (more common), retaining the top 10,000 common entities.
Next, we also adopt GPT3.5 to generate a hierarchical concept list for each entity. This list must include at least three concepts, where each subsequent concept is a broader or more abstract category of the preceding one. We avoid strict academic classifications to maintain the conceptual hierarchy's logic and clarity. For instance, the hierarchical concept list for the entity ``\texttt{Dog}'' is ``\texttt{[Household pets, Terrestrial mammals, Mammals]}'' instead of the strict biological hierarchy ``\texttt{[Canis, Canidae, Carnivora]}'', as starting with Canis makes it too easy to guess the \texttt{dog} based on prior knowledge.
After that, GPT3.5 is also used to provide the most similar entity for each entity in the dataset. For example, the entity most similar to a ``\texttt{dog}'' is identified as a ``\texttt{wolf}''.

\subsection{Difficulty Modes}

\textbf{Easy mode} is to provide a simple starting point which is the first concept from the hierarchical concept list of the target entity. It requires LLM participants to generate at most 20 yes-or-no questions to identify the target entity, which is the same as traditional Twenty Question game.
For example, in Fig.~\ref{fig:zhishang-framework}, the simple starting point for the target entity ``\texttt{Tiger}'' is ``\texttt{Pantherinae}'', which directly guides the LLMs to focus its inference within the realm of Pantherinae. The possible question is ``\emph{Does it have stripes?}'', etc.

\textbf{Medium mode} is to provide a harder starting point which is set as the second concept in our task from the hierarchical concept list of the target entity. It requires LLM participants to identify the target entity with a broader realm.
For example, in Fig.~\ref{fig:zhishang-framework}, the harder starting point for the target entity tiger is Felidae, which directly guides the LLMs to focus its inference within the realm of Felidae. The possible question is ``\emph{Is it large in size?}'' to confirm whether it belongs to Pantherinae.

\textbf{Hard mode} introduces a similar entity for generating wrong answers which is set as two in our task besides providing the simple starting point same as the easy mode.
It requires LLM participants to generate at most 20 yes-or-no questions to identify the target entity with misleading information.
For example, in Fig.~\ref{fig:zhishang-framework}, a possible question is ``\emph{Does it often engage in activities in trees?}'' and the correct answer is ``\emph{No}'' because the target entity tiger is usually active on the ground instead of the tree which indicates the similar entity ``\texttt{Leopard}''. LLM participants are expected not to navigate through this wrong answer and rethink the correct inference path.

\subsection{Evaluation Metrics}
We adopt accuracy and rounds as metrics to evaluate LLMs' performance in the proposed BrainKing.
Accuracy measures whether an LLM can infer the target entity within 20 rounds of questioning, with a successful guess scored as 1 and an unsuccessful one as 0. Rounds measures how many questions it takes to infer the entity. Once the entity is inferred, the game stops regardless of whether it is correct or incorrect, preventing the LLMs from exploring the full range of possible entities. If the entity is not inferred within Twenty Questions, we set its rounds as 30, as it is not feasible to continue BrainKing indefinitely.
We multiply the accuracy and the reciprocal of the rounds by 100 to obtain the accuracy win rate and rounds win rate, respectively. The average of the accuracy win rate and rounds win rate is then calculated to determine the total win rate.
The optimal score for accuracy win rate, rounds win rate, and total win rate is 100. The higher the value, the higher the win rate.

Moreover, we also introduce the ability to recognize confusion as an evaluation metric to determine whether an LLM can backtrack from misleading information in an answer and returns to the correct path of questioning. We use GPT4 to analyze the whole question-and-answer process of an LLM. If it exclusively follows the misleading direction without returning to the correct path, it is marked as 1; otherwise, it is marked as 0.

\section{Experiments}
In this section, we conduct extensive experiments to evaluate different LLMs' performance in the proposed BrainKing.

\subsection{Experimental Setups}
Our experiments are conducted on 8 Nvidia A100 GPUs, each with 80GB of memory, and we use PyTorch~\footnote{https://pytorch.org/} in Python~\footnote{https://www.python.org/}. We set the maximum sequence length for both input and output sequences to maximum 200 tokens. 
we use GPT4 to respond to the questions posed by each LLM.

\subsection{Datasets, Baselines and Metrics}
The baseline LLMs for this evaluation are BLOOM-7B~\citep{workshop2023bloom}
BLOOM-176B~\citep{workshop2023bloom},
Claude2~\citep{bai2022constitutional},
Falcon-7B~\citep{almazrouei2023falcon},
Falcon-180B~\citep{almazrouei2023falcon},
GPT3.5~\citep{brown2020language}, 
GPT4~\citep{openai2023gpt4},
LLaMA2-7B~\citep{touvron2023llama},
LLaMA2-70B~\citep{touvron2023llama},
Vicuna-7B~\citep{chiang2023vicuna}, and
Vicuna-33B~\citep{zheng2023judging}. The prompt for playing the BrainKing is shown in Table~\ref{tab:prompt}.

We recruit nine volunteers to participate in BrainKing. First, we select three entities, each representing a different mode of difficulty to test each volunteer's ability. Then, we rank their comprehensive scores from high to low. The top three scorers are assigned to hard mode, the middle three to medium mode, and the last three to easy mode. We then randomly distribute 1,000 entities, other than the initial three, among the nine volunteers and calculate their accuracy win rate, rounds win rate and  total win rate. The highest total win rate from each of the three modes are averaged again to obtain the Human performance. Volunteers participate on a voluntary basis without compensation.

\begin{table*}[]
\centering
\resizebox{\textwidth}{!}{%
\begin{tabular}{p{1.3cm}|p{25cm}}
\toprule
          Mode  & Prompt                                                                                                                                                                                                                                                                                                                                                                  \\ \midrule
Easy Mode   & Let's play a game of Twenty Questions. I have a hidden thing in mind and you need to guess it. This time, the thing belongs to \{a simple and straightforward starting point\}. You may only ask yes/no questions for information and I will only answer with yes or no, note that I will say yes when it is likely. When you make a final guess, say "Guess:(Your guess)". You're not allowed to make another guess after making the final guess. Now begin your first question, count the number of questions like this: "Q1:(Your question)"                                                                                                                                                                                                                         \\\midrule
Medium Mode & Let's play a game of Twenty Questions. I have a hidden thing in mind and you need to guess it. This time, the thing belongs to \{a harder starting point\}. You may only ask yes/no questions for information and I will only answer with yes or no, note that I will say yes when it is likely. When you make a final guess, say "Guess:(Your guess)". You're not allowed to make another guess after making the final guess. Now begin your first question, count the number of questions like this: "Q1:(Your question)"                                                                                                                                                                                                                                             \\\midrule
Hard Mode   & Let's play a game of Twenty Questions. I have a hidden thing in mind and you need to guess it. This time, the thing belongs to \{a simple and straightforward starting point\}. You may only ask yes/no questions for information and I will only answer with yes or no, note that I will say yes when it is likely. During the process, I will confuse the real item with another in the field, so some questions may not have been correctly answered, especially those more detailed. You may ask more questions and propose a new guess. When you make a final guess, say "Guess:(Your guess)". You're not allowed to make another guess after making the final guess. Now begin your first question, count the number of questions like this: "Q1:(Your question)" \\ \bottomrule
\end{tabular}%
}
\caption{The instruction prompts of three modes in guiding LLMs to generate questions.}
\label{tab:prompt}
\end{table*}

\subsection{Main results}





\emph{Question 1:Which LLM is the winner of the BrainKing? Answer 1: GPT4!}

\begin{figure}[!h]
  \centering
  \includegraphics[width=0.8\linewidth]{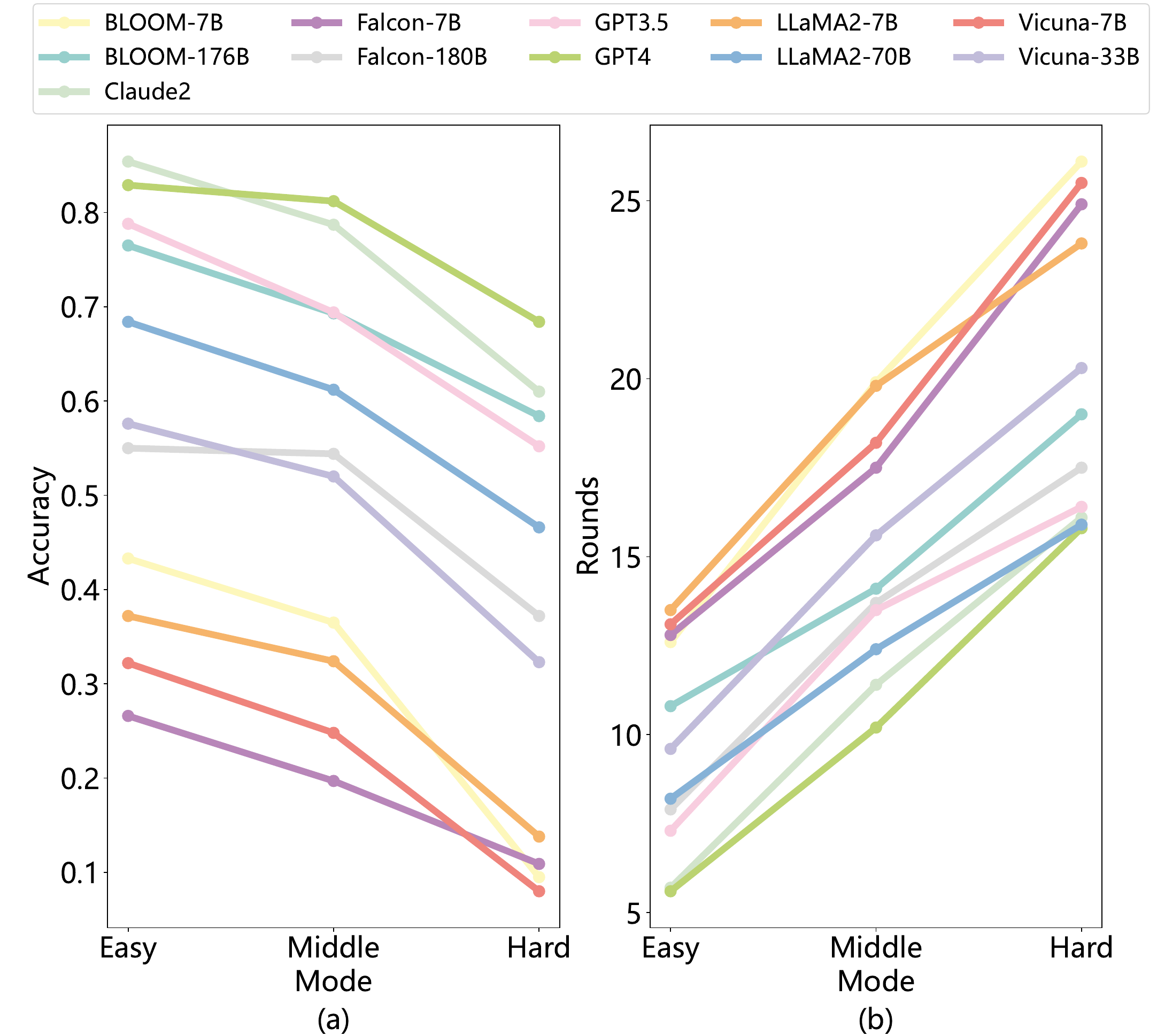}
  \caption{The comparison of the performance (accuracy and rounds) of different LLMs across three modes.}
  \label{fig:zhishang-exp1}
\end{figure}

The performance of different LLMs under three difficulty modes are shown in Table~\ref{tab:zhishang-exp1} and Fig.~\ref{fig:zhishang-exp1}. We observe that in the easy mode, Claude2 stands out with an accuracy of 85.4\% and an average of only 5.7 rounds. GPT4, GPT3.5, and BLOOM-176B also perform well with high accuracy above 70\% and few rounds below 11.
In the medium mode, GPT4 outperforms among all LLMs with an accuracy of 81.2\% and an average of 10.2 rounds, suggesting its powerfulness even with a harder starting point. Claude2, GPT3.5 and BLOOM-176B, despite a decrease in accuracy, still maintain around 70\%, showing good reasoning capabilities.
In the hard mode, further escalates the challenge by providing the same starting point as Easy Mode and introducing a similar entity that could cause confusion. Under these conditions, GPT4 also leads with an accuracy of 78.8\% and 15.8 rounds, followed by Claude2 and BLOOM-176B around 70\%. This suggests that GPT4 has strong robustness when facing with misleading information.
Across all modes, Falcon-7B and Vicuna-7B show poorer performance, especially in hard mode, which are below 20\%. This suggests that these LLMs are less capable of handling complex reasoning and resisting misleading information.

\begin{table}[]
\centering
\resizebox{0.47\textwidth}{!}{%
\begin{tabular}{@{}lcc|cc|cc@{}}
\toprule
             & \multicolumn{2}{c|}{Easy Mode} & \multicolumn{2}{c|}{Medium Mode} & \multicolumn{2}{c}{Hard Mode} \\ 
             & Accuracy       & Rounds       & Accuracy        & Rounds        & Accuracy       & Rounds       \\\midrule
BLOOM-7B     & 43.3           & 12.6         & 36.5            & 19.9          & 12.3           & 26.1         \\
BLOOM-176B   & 76.5           & 10.8         & 69.3            & 14.1          & 65.5           & 19.0         \\
Claude2     & \textbf{85.4}           & \underline{5.7}          & \underline{78.7}            & \underline{11.4}          & \underline{68.3}           & 16.1         \\
Falcon-7B    & 26.6           & 12.8         & 19.7            & 17.5          & 12.5           & 24.9         \\
Falcon-180B & 55.0           & 7.9          & 54.4            & 13.7          & 51.0           & 17.5         \\
GPT3.5       & 78.8           & 7.3          & 69.4            & 13.5          & 63.7           & 16.4         \\
GPT4         & \underline{82.9}           & \textbf{5.6}          & \textbf{81.2}            & \textbf{10.2}          & \textbf{78.8}           & \textbf{15.8}         \\
LLaMA2-7B   & 37.2           & 13.5         & 32.4            & 19.8          & 24.8           & 23.8         \\
LLaMA2-70B  & 68.4           & 8.2          & 61.2            & 12.4          & 55.7           & \underline{15.9}         \\
Vicuna-7B    & 32.2           & 13.1         & 24.8            & 18.2          & 17.4           & 25.5         \\
Vicuna-33B   & 57.6           & 9.6          & 52.0            & 15.6          & 47.8           & 20.3         \\ \bottomrule
\end{tabular}%
}
\caption{The accuracy and rounds of different LLMs in the proposed BrainKing benchmark.}
\label{tab:zhishang-exp1}
\end{table}

\begin{figure*}[!h]
  \centering
  \includegraphics[width=0.9\linewidth]{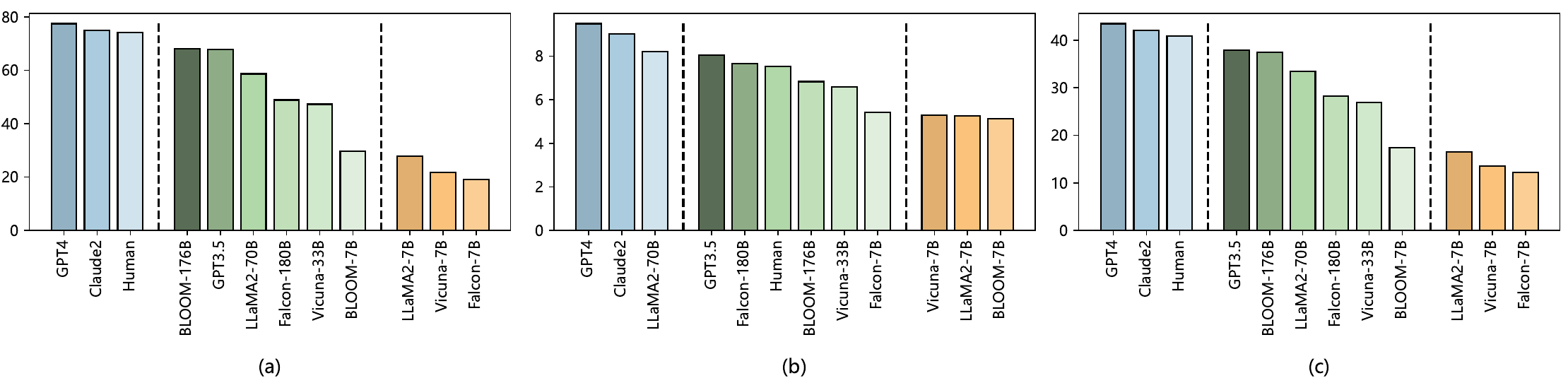}
  \caption{The win rate of different LLMs in the proposed BrainKing benchmark.}
  \label{fig:zhishang-exp2}
\end{figure*}

We calculate the win rates for accuracy and rounds, as well as the total win rate, and rank them accordingly. These rankings are displayed in Table~\ref{tab:zhishang-exp2}. 
Figure~\ref{fig:zhishang-exp2} displays three tiers of performance, from strongest to weakest, across the three win rates.
We observe that GPT4 consistently performs best across accuracy win rate, rounds win rate, and total win rate, demonstrating its strong capabilities. 
Claude2 follows closely, ranking second with a 75\% accuracy win rate and a 9.0 rounds win rate, and a total win rate of 42.0\%. This indicates that Claude2 is a very powerful model, effectively competing with GPT4.
Human also performs well, ranking in the top three for accuracy win rate and total win rate, but fall to sixth place in rounds win rate, which may suggest that while human participants can infer the entity but not much efficient compared with GPT4 and Claude2.
GPT3.5, BLOOM-176B, LLaMA2-70B and Falcon-180B have moderate performances, while BLOOM-7B, Vicuna-33B, and Vicuna-7B lag behind, especially in the accuracy win rate.
LLaMA2-7B and Falcon-7B are at the bottom, ranking low in all metrics, particularly in the total win rate, which have more room for optimization.

\emph{Question 2: Are accuracy and rounds in a completely inverse relationship? Answer 2: No, just inversely correlated!}

\begin{figure*}[!h]
  \centering
  \includegraphics[width=0.9\linewidth]{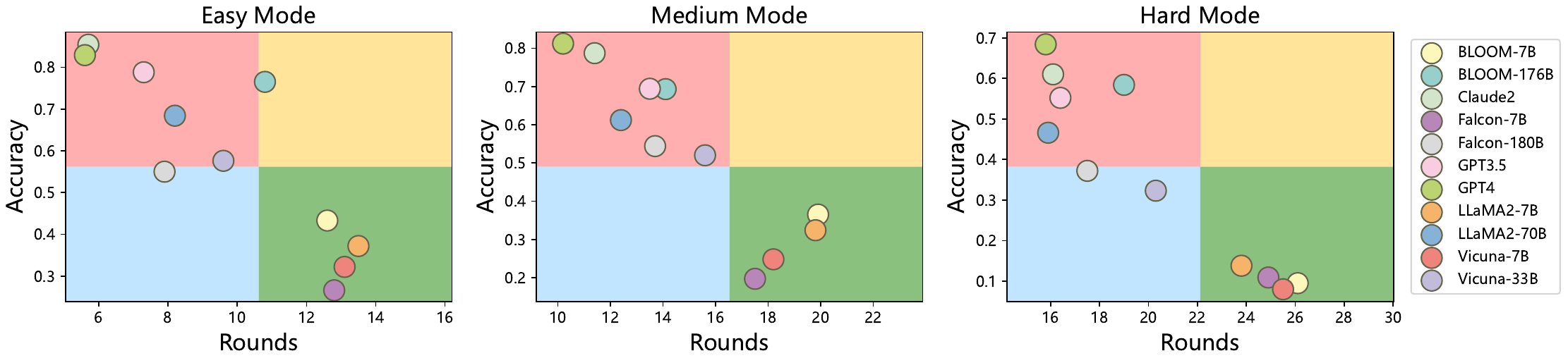}
  \caption{The relationship between accuracy and rounds across three modes of different LLMs.}
  \label{fig:zhishang-exp3}
\end{figure*}

We also analyze the relationship between accuracy and rounds in three modes as shown in Fig.~\ref{fig:zhishang-exp3}. We find that the accuracy and the rounds do not show a strict inverse relationship. In the easy mode, most LLMs achieve high accuracy with fewer rounds. However, some LLMs, such as those with 7B size, still show low accuracy after many rounds, which suggests that not every round of answers is effectively used for reasoning.
In the medium and hard mode, there's a more noticeable trend of accuracy decreasing as the number of rounds increases, which suggests that LLMs need more information to make accurate reasoning and are more susceptible to being misled.
Therefore, the relationship between accuracy and rounds is not strictly inverse but a more complex interplay where many factors can affect the outcome.

\begin{table}[]
\centering
\resizebox{0.45\textwidth}{!}{%
\begin{tabular}{@{}lcc|cc|cc@{}}
\toprule
            & \multicolumn{2}{c|}{Accuracy} & \multicolumn{2}{c|}{Rounds} & \multicolumn{2}{c}{Total} \\ 
            & Win Rate        & Rank       & Win Rate       & Rank      & Win Rate       & Rank       \\\midrule
GPT4        & \textbf{77.5}            & 1          & \textbf{9.5}            & 1         & \textbf{43.5}           & 1          \\
Claude2     & \underline{75.0}              & 2          & \underline{9.0}            & 2         & \underline{42.0}           & 2          \\
Human       & 74.2            & 3          & 7.5            & 6         & 40.9           & 3          \\
GPT3.5      & 67.8            & 5          & 8.1            & 4         & 37.9           & 4          \\
BLOOM-176B & 68.1            & 4          & 6.8            & 7         & 37.5           & 5          \\
LLaMA2-70B  & 58.7            & 6          & 8.2            & 3         & 33.5           & 6          \\
Falcon-180B & 48.9            & 7          & 7.7            & 5         & 28.3           & 7          \\
Vicuna-33B & 47.3            & 8          & 6.6            & 8         & 26.9           & 8          \\
BLOOM-7B   & 29.8            & 9          & 5.1            & 12        & 17.5           & 9          \\
LLaMA2-7B   & 27.8            & 10         & 5.3            & 11        & 16.5           & 10         \\
Vicuna-7B  & 21.7            & 11         & 5.3            & 10        & 13.5           & 11         \\
Falcon-7B   & 19.1            & 12         & 5.4            & 9         & 12.3           & 12         \\ \bottomrule
\end{tabular}%
}
\caption{The win rate of different LLMs in the proposed BrainKing benchmark.}
\label{tab:zhishang-exp2}
\end{table}

\emph{Question 3: Does the difficulty of starting points significantly affect LLMs' performance? Answer 3: More significant for weaker LLMs!}

\begin{figure}[!h]
  \centering
  \includegraphics[width=0.8\linewidth]{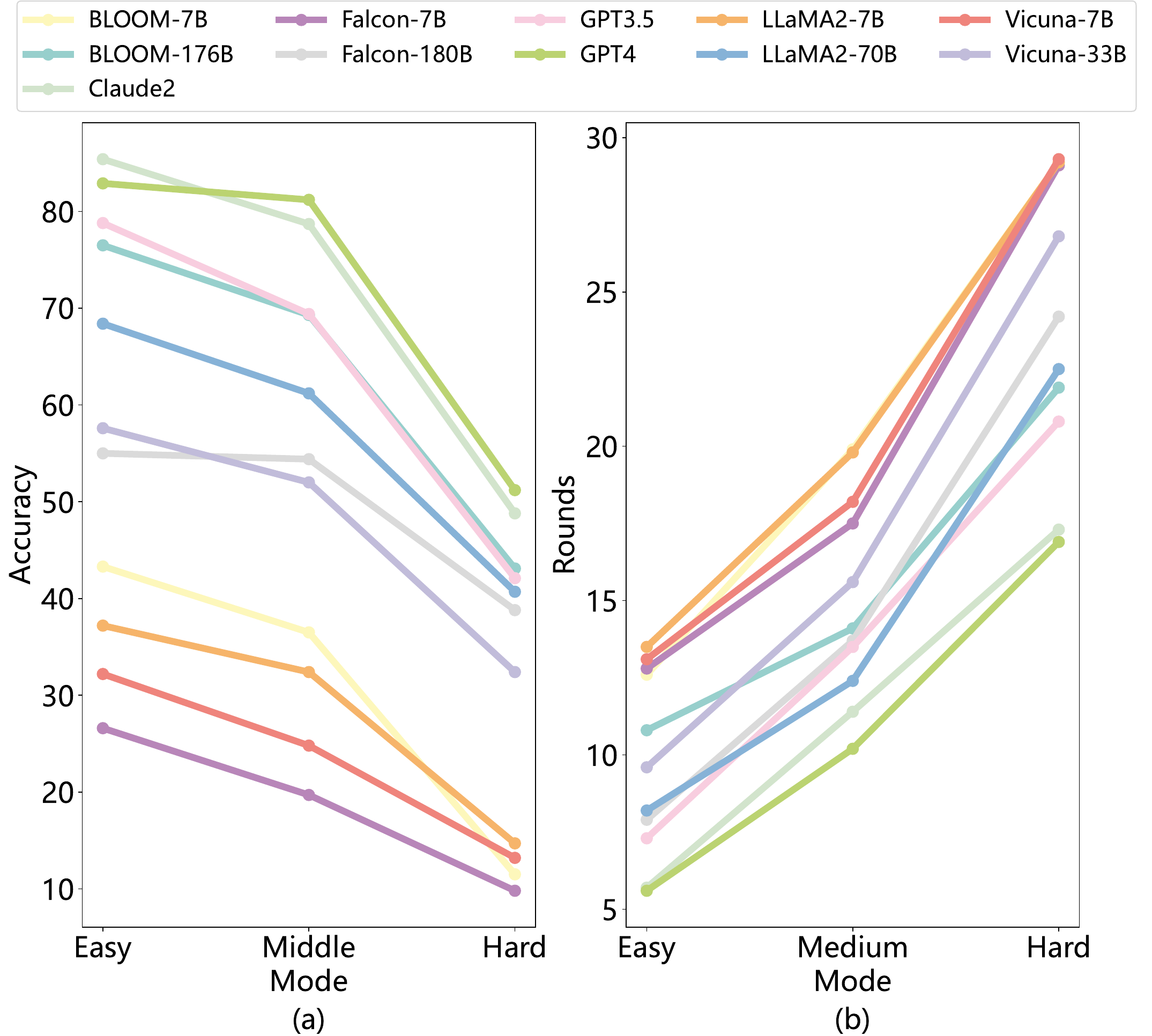}
  \caption{The performance (accuracy and rounds) of different LLMs with different difficulties of starting points.}
  \label{fig:zhishang-exp4}
\end{figure}

We also analyze the change in model performance with the increase in the difficulty of starting points as shown in Table~\ref{tab:zhishang-exp3-1} and Fig.~\ref{fig:zhishang-exp4}. As the starting point difficulty goes up, most LLMs tend to show a significant decrease in accuracy. At easy starting points, Claude2 and GPT4 demonstrate higher accuracy, and although their accuracy drops at hard starting points, they still maintain relatively high levels.
Additionally, as the difficulty increases, the number of rounds needed tends to go up. For example, BLOOM-176B requires 10.8 rounds at the easy starting point, which increases to 21.9 rounds at the hard starting point.
Overall, the difficulty of the starting point has a clear impact on LLMs' performance. Strong LLMs still demonstrate their reasoning capabilities even at hard starting points, while LLMs with weaker performance are more significantly affected in both accuracy and number of rounds.

\begin{table}[]
\centering
\resizebox{0.47\textwidth}{!}{%
\begin{tabular}{@{}lcc|cc|cc@{}}
\toprule
             & \multicolumn{2}{c|}{Easy Starting Point} & \multicolumn{2}{c|}{Medium Starting Point} & \multicolumn{2}{c}{Hard Starting Point} \\ 
             & Accuracy            & Rounds            & Accuracy             & Rounds             & Accuracy            & Rounds            \\\midrule
BLOOM-7B     & 43.3                & 12.6              & 36.5                 & 19.9               & 11.5                & 29.2              \\
BLOOM-176B   & 76.5                & 10.8              & 69.3                 & 14.1               & 43.1                & 21.9              \\
Claude2     & \textbf{85.4}                & \underline{5.7}               & \underline{78.7}                 & \underline{11.4}               & \underline{48.8}                & \underline{17.3}              \\
Falcon-7B    & 26.6                & 12.8              & 19.7                 & 17.5               & 9.8                 & 29.1              \\
Falcon-180B & 55.0                  & 7.9               & 54.4                 & 13.7               & 38.8                & 24.2              \\
GPT3.5       & 78.8                & 7.3               & 69.4                 & 13.5               & 42.1                & 20.8              \\
GPT4         & \underline{82.9}                & \textbf{5.6}               & \textbf{81.2}                 & \textbf{10.2}               & \textbf{51.2}                & \textbf{16.9}              \\
LLaMA2-7B   & 37.2                & 13.5              & 32.4                 & 19.8               & 14.7                & 29.2              \\
LLaMA2-70B  & 68.4                & 8.2               & 61.2                 & 12.4               & 40.7                & 22.5              \\
Vicuna-7B    & 32.2                & 13.1              & 24.8                 & 18.2               & 13.2                & 29.3              \\
Vicuna-33B   & 57.6                & 9.6               & 52.0                   & 15.6               & 32.4                & 26.8              \\ \bottomrule
\end{tabular}%
}
\caption{The performance (accuracy and rounds) of different LLMs with different difficulties of starting points.}
\label{tab:zhishang-exp3-1}
\end{table}

\emph{Question 4: Does the number of wrong answers significantly affect LLMs' performance? Answer 4: Yes, most LLMs are nearing the upper limit of rounds!}

\begin{figure}[!h]
  \centering
  \includegraphics[width=0.8\linewidth]{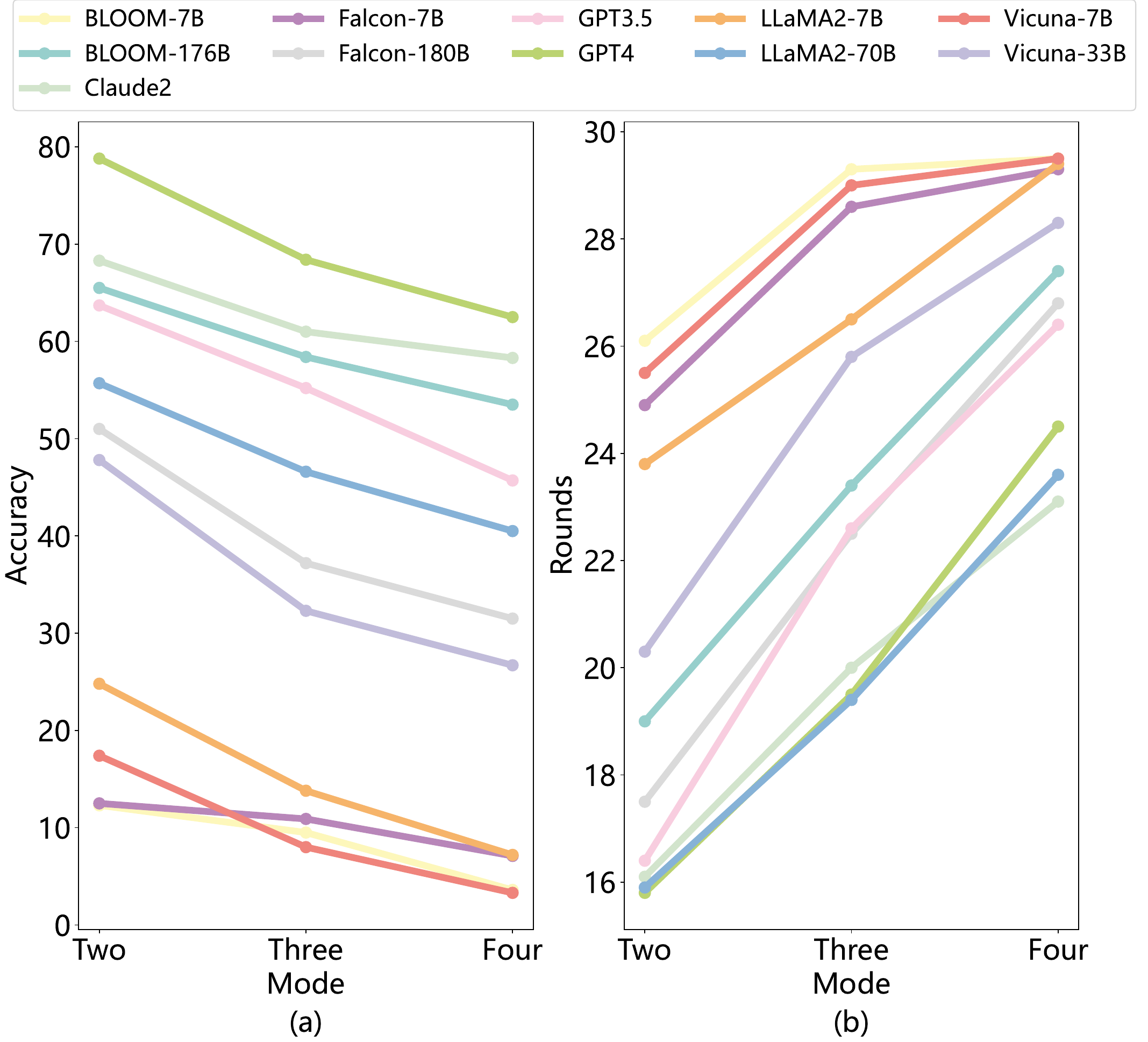}
  \caption{The performance (accuracy and rounds) of different LLMs with different number of wrong answers.}
  \label{fig:zhishang-exp5}
\end{figure}

We also analyze whether the number of wrong answers significantly affect LLMs' performance as shown in Table~\ref{tab:zhishang-exp3-2} and Fig.~\ref{fig:zhishang-exp5}. We observe that with an increase in the number of wrong answers, the accuracy of all LLMs generally decreases, while the number of rounds needed increases or remains relatively stable. The lack of change of rounds is mainly because the limit of rounds (that is 20) has been reached. For example, GPT4 has an accuracy rate of 78.8\% with just two wrong answers, but this drops to 62.5\% with four wrong answers.
Particularly for poorly performing LLMs, such as Falcon-7B and Vicuna-7B, their accuracy dramatically drop almost to the point of indistinction as the number of wrong answers increases, demonstrating their vulnerability in processing misleading information.
In terms of rounds, most models either need more rounds to answer or show no significant change when the number of wrong answers increases. For instance, BLOOM-176B needs 19.0 rounds with two wrong answers, but this number increases to 27.4 rounds with four wrong answers.

\begin{table}[]
\centering
\resizebox{0.47\textwidth}{!}{%
\begin{tabular}{@{}lcc|cc|cc@{}}
\toprule
            & \multicolumn{2}{c|}{Two Wrong Answers} & \multicolumn{2}{c|}{Three Wrong Answers} & \multicolumn{2}{c}{Four Wrong Answers} \\ 
            & Accuracy          & Rounds            & Accuracy           & Rounds             & Accuracy           & Rounds            \\\midrule
BLOOM-7B    & 12.3              & 26.1              & 9.5                & 29.3               & 3.6                & 29.5              \\
BLOOM-176B  & 65.5              & 19.0              & 58.4               & 23.4               & 53.5               & 27.4              \\
Claude2     & \underline{68.3}        & 16.1              & \underline{61.0}         & \underline{20.0}         & \underline{58.3}         & \textbf{23.1}     \\
Falcon-7B   & 12.5              & 24.9              & 10.9               & 28.6               & 7.1                & 29.3              \\
Falcon-180B & 51.0              & 17.5              & 37.2               & 22.5               & 31.5               & 26.8              \\
GPT3.5      & 63.7              & 16.4              & 55.2               & 22.6               & 45.7               & 26.4              \\
GPT4        & \textbf{78.8}     & \textbf{15.8}     & \textbf{68.4}      & 19.5               & \textbf{62.5}      & 24.5              \\
LLaMA2-7B   & 24.8              & 23.8              & 13.8               & 26.5               & 7.2                & 29.4              \\
LLaMA2-70B  & 55.7              & \underline{15.9}        & 46.6               & \textbf{19.4}      & 40.5               & \underline{23.6}        \\
Vicuna-7B   & 17.4              & 25.5              & 8.0                & 29.0               & 3.3                & 29.5              \\
Vicuna-33B  & 47.8              & 20.3              & 32.3               & 25.8               & 26.7               & 28.3              \\ \bottomrule
\end{tabular}%
}
\caption{The performance (accuracy and rounds) of different LLMs with different number of wrong answers.}
\label{tab:zhishang-exp3-2}
\end{table}

\emph{Question 5: In the hard mode, is there a strict proportional relationship between accuracy and the rethink ability? Answer 5: No, they are only positively correlated and the rating of rethink ability is generally higher than accuracy!}



\begin{figure}[!h]
  \centering
  \includegraphics[width=\linewidth]{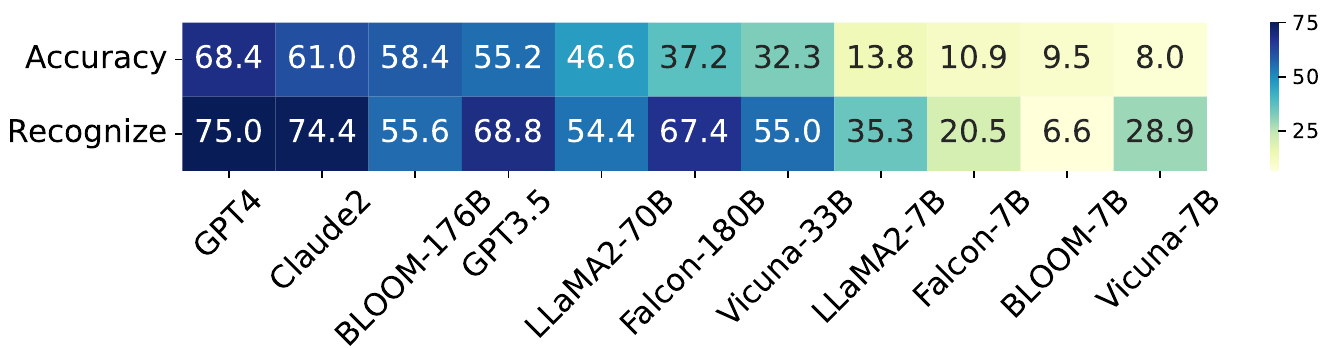}
  \caption{The Pearson correlation coefficient that normalized to a 1-100 scale between accuracy and rethink capabilities of different LLMs.}
  \label{fig:zhishang-exp6}
\end{figure}

We adopt Pearson correlation coefficient that normalized to a 1-100 scale to analyze the relationship between the accuracy of LLMs and their rethink abilities that backtrack from misleading information as shown in Fig.~\ref{fig:zhishang-exp6}. The closer the colors, the higher the correlation.
We find that most LLMs show similar shades, indicating a positive correlation but not a strict proportional relationship between their accuracy and their rethink abilities to detect the confusion.
Specifically, GPT4 and Claude2 score high in both accuracy and rethink abilities, with a deeper blue color indicating a strong relationship between the two performances. In contrast, LLaMA2-7B, Falcon-7B, and BLOOM-7B score lower in both metrics and show lighter colors, suggesting weaker performance and lower correlation.
Furthermore, the scores for the ability to rethink are generally higher than those for accuracy, which implies that even though a language model can backtrack when faced with misleading information, it does not always result in high accuracy in the final judgment. This might be because language models still struggle to effectively utilize this information, particularly in cases where the information is ambiguous.

\subsection{Case study}

\begin{figure*}[!h]
  \centering
  \includegraphics[width=0.9\linewidth]{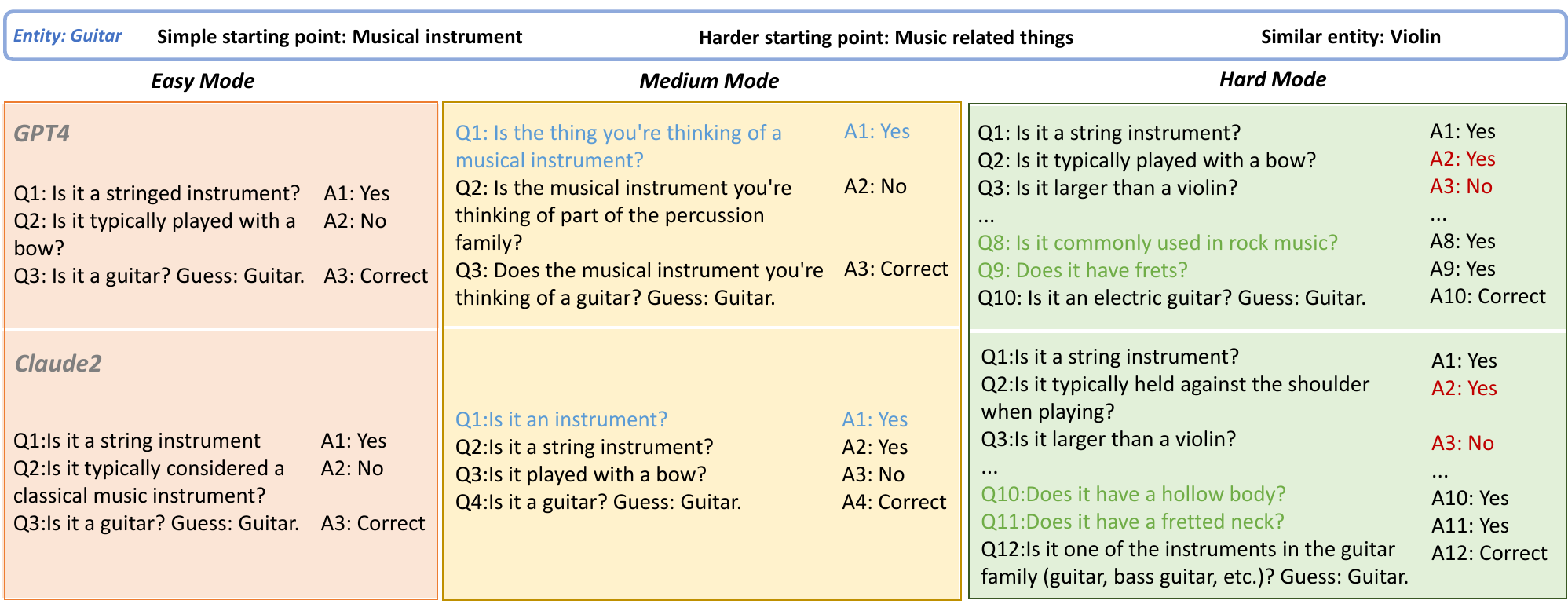}
  \caption{Questions posed by the top2 LLMs, i.e. GPT4 and Claude2, among three modes for a given entity.}
  \label{fig:zhishang-case}
\end{figure*}

\begin{figure}[!h]
  \centering
  \includegraphics[width=0.9\linewidth]{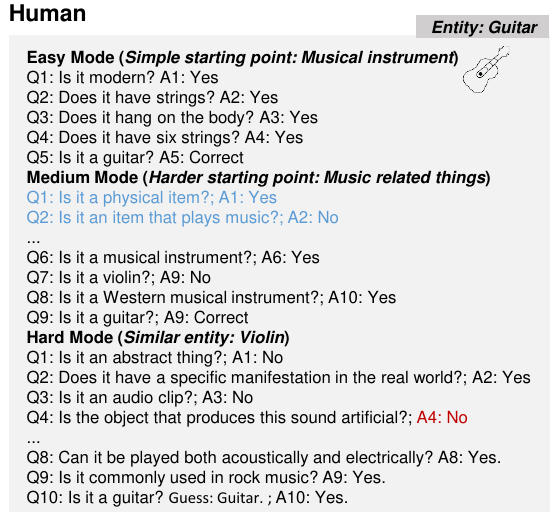}
  \caption{Questions posed by human among three modes for a given entity.}
  \label{fig:zhishang-human}
\end{figure}

We show a good running example with questions generated by the top two LLMs (i.e. GPT4 and Claude2) and human in Fig.~\ref{fig:zhishang-case} and Fig.~\ref{fig:zhishang-human}, respectively. More cases are shown in Fig.~\ref{fig:zhishang-case1}, Fig.~\ref{fig:zhishang-case2}.

In Fig.~\ref{fig:zhishang-case}, we find that both GPT4 and Claude2 make correct reasoning with just three questions in the easy mode. GPT4 focuses on determining the type of musical instrument and excluding percussion instruments, while Claude2 directly asks if it is a string instrument and if it belongs to classical music instruments. This shows that both are efficient in identifying with basic questions. 
In the medium mode with a harder starting point, GPT4 needs ten questions, while Claude2 only needs four. GPT4 starts by confirming it is a string instrument and narrows down to a guitar, including questions about whether it is played with a bow and if it is common in rock music. In contrast, Claude2's questions are more direct, quickly moving from whether it is a string instrument and played with a bow to identifying it as a guitar. This suggests that Claude2 is slightly more efficient and accurate in a more complex situation.
In the hard mode, both GPT4 and Claude2 show the ability to narrow down gradually to identify the guitar. GPT4 takes ten questions, starting with confirming it is a string instrument and then asking if it is common in rock music and if it has frets to exclude the violin. Claude2 needs twelve questions, also starting with confirming it is a string instrument, but focusing more on physical characteristics like size and whether it has a hollow body, eventually asking if it is part of the guitar family. In this situation, GPT4 shows higher efficiency, although both are accurate.

Overall, the differences between them may stem from their specific ways of processing information and decision-making.
However, there are also some bad cases, including wrong logic reasoning that ask same questions which have been answered before, not following instructions that asks non-yes-or-no questions or self-questioning as shown in Fig.~\ref{fig:zhishang-case2}.

\section{Related Work}

\subsection{Gaming abilities of LLMs}
Recent research shows the capabilities of LLMs in various gaming scenarios. For example, 
\citet{zhao2023solving} evaluate LLMs in solving and creating puzzles for the NPR Sunday Puzzle game show;
\citet{jiang2023Brainteaser} introduce BRAINTEASER to assess lateral thinking in LLMs; 
\citet{lorè2023strategic} investigate strategic decision-making in LLMs through game theory;
\citet{egan2022play} propose new summary evaluation metrics using the Shannon Game; 
\citet{brookins2023playing} study LLM preferences in strategic games like the dictator game and the prisoner's dilemma;
\citet{ogara2023hoodwinked} explore deception and lie detection in LLMs through the game Hoodwinked;
\citet{liga2023testing} investigate the spatial reasoning of LLMs in the game of tic-tac-toe. 
%
Inspired but different from the previous work, our research evaluates the information processing and problem-solving capability of LLMs under the incomplete information scenario including but not limited to their knowledge retrieval, misleading information recognition capabilities through one simple game.

\subsection{Evaluation for LLMs' Capabilities}
Recent research has extensively explored the capabilities of LLMs across various domains~\citep{chen2023hadamard,chen2023hallucination,ren2024survey}.
For example,
~\citet{ziems2023large} demonstrate that LLMs can significantly contribute to Computational Social Science by classifying and explaining social phenomena; 
~\citet{zheng2023judging} explore the use of LLMs as judges to evaluate chat assistants, introducing benchmarks like MT-bench and Chatbot Arena; 
~\citet{zhong2023chatgpt} focus on comparing the understanding ability of ChatGPT with fine-tuned BERT-style models using the GLUE benchmark;
~\citet{laskar2023systematic} present a comprehensive evaluation of ChatGPT on diverse academic datasets, including question-answering, text summarization, code generation, commonsense reasoning;
~\citet{valmeekam2023planbench} introduce PlanBench, a benchmark for evaluating LLMs on planning and reasoning;
~\citet{del2023true} introduce a benchmark consisting of long-form mystery narratives in assessing LLMs' advanced reasoning abilities;
~\citet{sawada2023arb} propose a benchmark containing advanced reasoning problems across multiple domains to evaluate the advanced reasoning capabilities of LLMs;
~\citet{valmeekam2023planning} investigate the planning abilities of LLMs in commonsense tasks and as heuristic guidance for other agents;
While the aforementioned studies design different benchmarks in evaluating LLMs' capabilities, there is lack of a benchmark for evaluating the information processing and problem-solving capability of LLMs under incomplete information scenarios.

\section{Conclusions and Future Work}
In conclusion, our study highlights the importance of a multifaceted approach to evaluating the information processing and problem-solving capability of LLMs under incomplete information scenarios. The BrainKing game, as a novel benchmark, successfully challenges LLMs in various cognitive aspects, from basic knowledge retrieval to complex reasoning and confusion recognition. Our findings offer a detailed understanding of the strengths and limitations of current LLMs, underscoring the need for LLMs to not only process information accurately but also navigate through misleading information effectively.
For future work, there is potential to extend the complexity and scope of the BrainKing, introducing more challenging scenarios to better mimic real-world conditions. 

\section*{Limitations}
There are a few key limitations. First, the Twenty Questions dataset used may not fully represent all possible types of entities. Moreover, selecting only the top 10,000 most common entities might limit the breadth and depth of the assessment.
Second, in the hard mode of the game, LLMs are required to correctly identify the target entity based on potentially misleading information. This might be an overly challenging demand, especially for LLMs not yet fully trained to handle such complex tasks.
Third, determining whether an LLM can recover from misleading information and return to the correct line of questioning can be a complex process. It involves a detailed analysis of the entire question-and-answer session, which could add to the complexity and subjectivity of the evaluation.


\section*{Acknowledgements}
This work is supported by
Science and Technology Commission of Shanghai Municipality Grant (No. 22511105902), Shanghai Municipal Science and Technology Major Project (No.2021SHZDZX0103), the National Natural Science Foundation of China (No.62072323), Shanghai Science and Technology Innovation Action Plan (No. 22511104700), and the Zhejiang Lab Open Research Project (NO. K2022NB0AB04).

\bibliography{anthology,main}

\appendix

\begin{figure*}[!h]
  \centering
  \includegraphics[width=\linewidth]{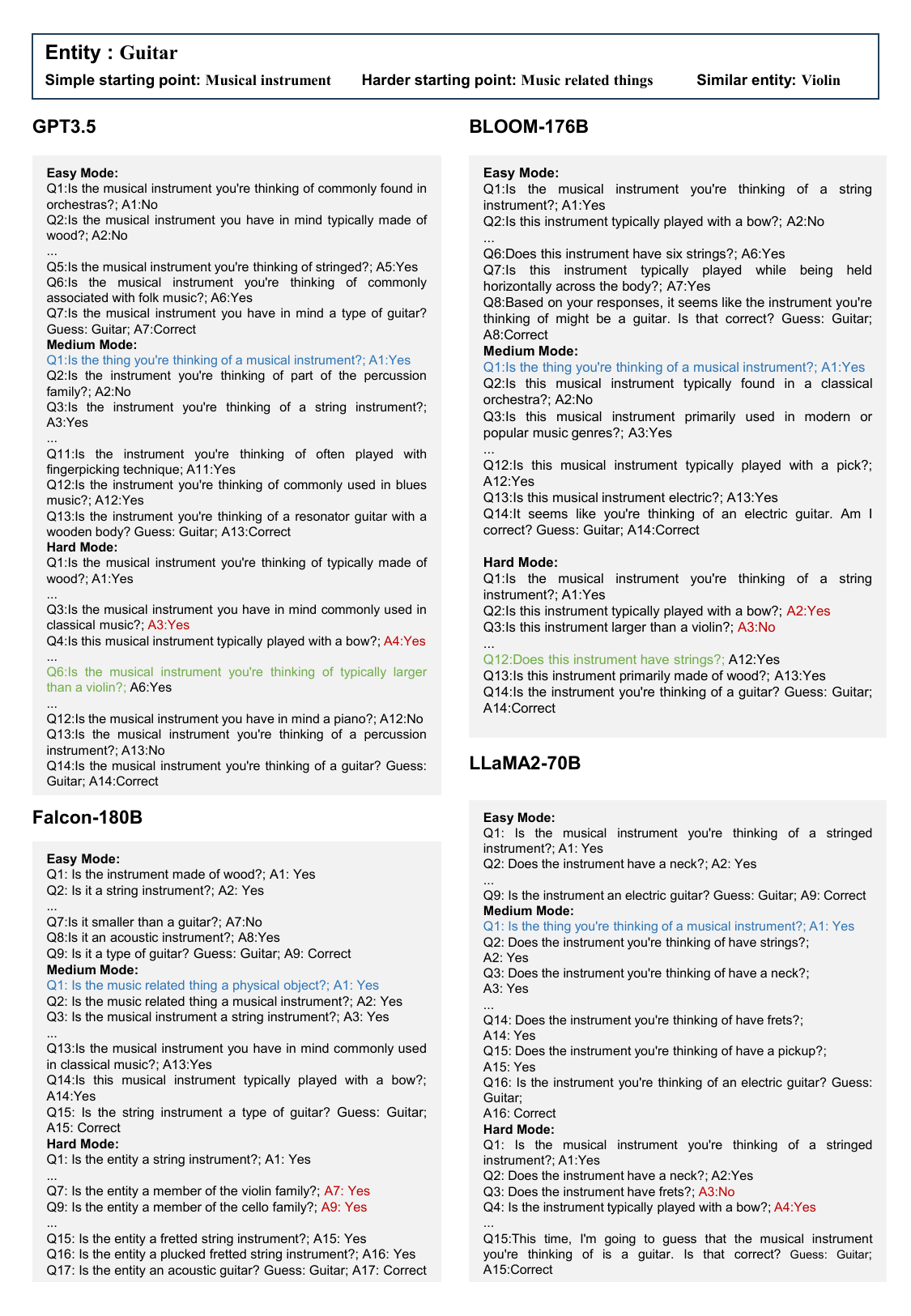}
  \caption{Questions posed by different LLMs among three modes for a given entity.}
  \label{fig:zhishang-case1}
\end{figure*}

\begin{figure*}[!h]
  \centering
  \includegraphics[width=\linewidth]{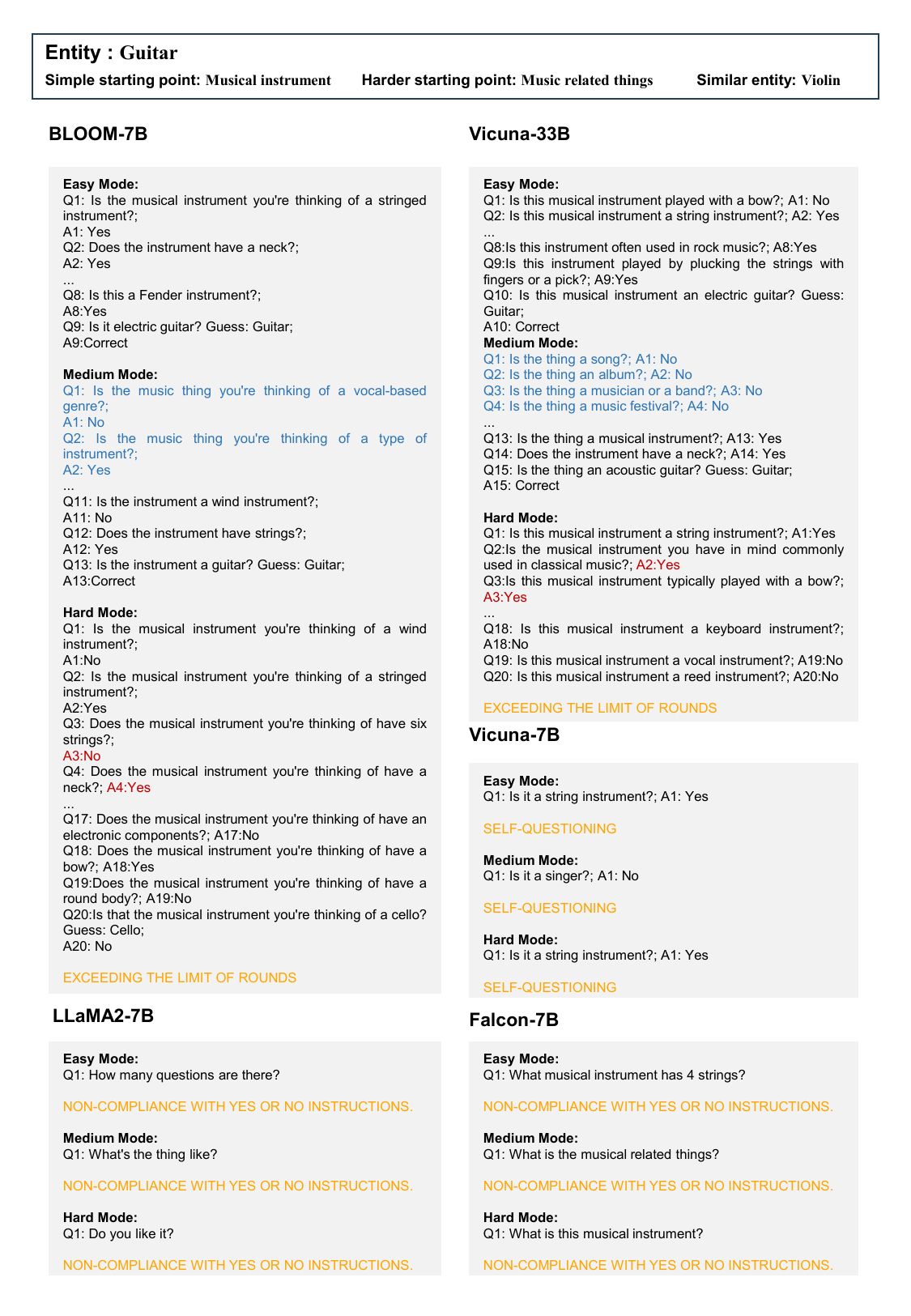}
  \caption{Questions posed by different LLMs among three modes for a given entity.}
  \label{fig:zhishang-case2}
\end{figure*}

\end{document}